\documentclass{article}

\usepackage{microtype}
\usepackage{graphicx}
\usepackage{subfigure}
\usepackage{xspace}
\usepackage{booktabs} 

\usepackage{hyperref}
\newcommand{\namemodel}{VLM-CaR\xspace}

\DeclareFixedFont{\ttb}{T1}{txtt}{bx}{n}{12} 
\DeclareFixedFont{\ttm}{T1}{txtt}{m}{n}{12}  

\usepackage{color}
\definecolor{deepblue}{rgb}{0,0,0.5}
\definecolor{deepred}{rgb}{0.6,0,0}
\definecolor{deepgreen}{rgb}{0,0.5,0}

\usepackage{enumitem}

\usepackage{listings}

\newcommand\pythonstyle{\lstset{
language=Python,
basicstyle=\ttm,
morekeywords={self},              
keywordstyle=\ttb\color{deepblue},
emph={MyClass,__init__},          
emphstyle=\ttb\color{deepred},    
stringstyle=\color{deepgreen},
frame=tb,                         
showstringspaces=false
}}

\lstnewenvironment{python}[1][]
{
\pythonstyle
\lstset{#1}
}
{}

\newcommand\pythoninline[1]{{\pythonstyle\lstinline!#1!}}

\usepackage{tikz}

\usepackage[accepted]{icml2024}

\usepackage{amsmath}
\usepackage{amssymb}
\usepackage{mathtools}
\usepackage{amsthm}
\usepackage{float}
  \usepackage{placeins}

\usepackage[capitalize,noabbrev]{cleveref}

\theoremstyle{plain}

\theoremstyle{definition}

\theoremstyle{remark}

\usepackage[textsize=tiny]{todonotes}

\newcommand{\gpt}[1]{{\color{green} #1}}
\newcommand{\user}[1]{{\color{blue} #1}}

\icmltitlerunning{Code as Reward: Empowering Reinforcement Learning with VLMs}

\begin{document}

\twocolumn[
\icmltitle{Code as Reward: Empowering Reinforcement Learning with VLMs}




\begin{icmlauthorlist}
\icmlauthor{David Venuto}{yyy,comp}
\icmlauthor{Sami Nur Islam}{comp}
\icmlauthor{Martin Klissarov}{yyy,comp}
\icmlauthor{Doina Precup}{yyy,comp,ttt}
\icmlauthor{Sherry Yang}{ttt,brk}
\icmlauthor{Ankit Anand}{ttt}

\end{icmlauthorlist}

\icmlaffiliation{yyy}{Mila}
\icmlaffiliation{comp}{McGill University}
\icmlaffiliation{ttt}{Google DeepMind}
\icmlaffiliation{brk}{University of California, Berkeley}

\icmlcorrespondingauthor{David Venuto}{david.venuto@mail.mcgill.ca}

\icmlkeywords{Machine Learning, ICML}

\vskip 0.3in
]



\printAffiliationsAndNotice{} 

\begin{abstract}
Pre-trained Vision-Language Models (VLMs) are  able to understand visual concepts, describe and decompose complex tasks into sub-tasks, and provide feedback on task completion. In this paper, we aim to leverage these capabilities to support the training of reinforcement learning (RL) agents. In principle, VLMs are well suited for this purpose, as they can naturally analyze image-based observations and provide feedback (reward) on learning progress. However, inference in VLMs is computationally expensive, so querying them frequently to compute rewards would significantly slowdown the training of an RL agent. To address this challenge, we propose a framework named Code as Reward (\namemodel). \namemodel produces dense reward functions from VLMs through code generation, thereby significantly reducing the computational burden of querying the VLM directly. 
We show that the dense rewards generated through our approach are very accurate across a diverse set of discrete and continuous environments,  and can be more effective in training RL policies than the original sparse environment rewards.
\end{abstract}

\section{Introduction}
\label{intro}

One of the important steps in deploying reinforcement learning (RL) agents in  real-world problems with RL is defining an appropriate reward function, which both captures the goal of the problem and allows an RL agent to learn efficiently. The reward function is often defined and refined by an environment designer, with significant effort. In this paper, we explore the use of  large Vision-Language models (VLMs), which have recently become publicly available, for this purpose. VLMs are capable of learning intricate relationships between images and text. Integrated with large language models (LLMs), VLMs have been applied across a wide range of multi-modal tasks, from visual question answering~\citep{zhu2023minigpt} to robotics~\citep{brohan2023rt}. 
LLMs have also been used to craft better exploration strategies for RL agents, e.g. \citep{klissarov2023motif, yu2023language} 
or to produce programs that represent policies~\citep{liang2023code} or skills \citep{wang2023voyager}. 
Since pre-trained VLMs can accurately describe tasks and generate sub-tasks from image inputs, they are perfect candidates for enhancing RL agents that work with image-based or multi-modal observations. 

 In this paper, we aim to leverage VLMs for computing rewards from images, which can then be used to train RL agents.
The most direct approach to achieve this goal is to directly finetune a VLM to be a reward model, as in~\citep{yang2023learning,du2023video}. However, directly querying a VLM to obtain reward has a few disadvantages. First, it is computationally inefficient due to the sheer size of VLMs. Second, the stochastic output from a VLM can be unreliable. Third, rewards directly computed by a VLM are generally not interpretable nor verifiable. To address these problems, we present a framework called {Code as Reward (\namemodel)} which leverages pretrained VLMs to obtain efficiently computable, reliable, and verifiable reward functions. Our approach is to first prompt the VLM to describe the task/sub-tasks solely from an initial and goal frame of an environment. Given these high-level language descriptions, we use the VLM to generate a set of executable computer programs, one for each proposed sub-task, which check for successful sub-task completion.
To verify that the task decomposition and reward function programs are indeed correct, we propose an automated \emph{code verifier}, which uses a handful of expert trajectories and random trajectories to validate the obtained rewards. If the produced solution is correct, expert trajectories should achieve high reward, whereas random policies should achieve low reward. Once the generated sub-task and reward programs pass the verification check, we use them as reward functions  to train RL agents.  This approach can be viewed as an implementation of the  option framework in RL~\citep{SUTTON1999181}, where the VLM is used to produce the reward functions for training each option, as well as for the policy over options, and the rewards are represented programatically.
We show that we can successfully train RL agents using the rewards provided by \namemodel across a variety of grid-based discrete action environments and continuous robotic control environments.  Our approach has the additional benefit of providing dense rewards, which mitigates the notorious difficulty of training RL agents in sparse reward settings~\citep{hare2019dealing}.  

Our contributions can be summarized as follows:

\begin{itemize}[leftmargin=*,topsep=0pt,itemsep=0pt]
\item We propose a framework for prompting VLMs to generate sub-tasks and reward functions, represented as code, and discuss the relationship of this approach to the options framework.
\item We propose a reliable approach for verifying and refining the generated programs, by using a small amount of expert trajectories.
\item We show that the proposed approach enables effective training of RL policies, by providing dense rewards for tasks where the reward is either sparse or not available (which is often the case in real-world problems).
\end{itemize}

\section{Related Work}

\subsection{Vision-Language Models}
VLMs extract embeddings from both visual and text inputs. Convolutional layers  are used to encode visual features, while transformers are used to process text data; a cross-modal attention mechanism is used to establish connections between the  embeddings of the image and text inputs. During this process, the model learns to attend to relevant parts of the image while processing the text and vice versa.

Research has explored the capabilities of VLMs to understand complex relationships between images and text. VLMs like CLIP (Contrastive Language-Image Pre-training) \citep{radford2021learning} and UNITER (Unified Pre-trained Image-Text Transformer) \citep{10.1007/978-3-030-58577-8_7} have demonstrated impressive performance in various tasks involving both vision and language, such as scene description.  Larger models such as GPT-4 \citep{openai2023gpt4} have recently become available and showcase noteworthy performance across breadth of tasks, as well as with increasing task complexity. \namemodel builds upon these advancements by extending the utility of VLMs to reinforcement learning, specifically by providing reward design in an interpretable and automated manner.

\subsection{VLMs as Reward Models}


VLMs are particularly useful for crafting rewards functions in RL, as they directly work at the intersection of visual observations and language descriptions. OpenAI's CLIP model \citep{radford2021learning} has been  used to derive dense reward functions, by taking the cosine similarity between the embeddings of a goal description in language and a visual representation of the current state (or possibly a sequence of states). MineDojo \citep{Fan2022MineDojoBO} builds on the CLIP model and fine-tunes it on thousands of hours of YouTube videos showcasing human game-play in MineCraft \citep{Kanervisto2022MineRLD2}. The result is the MineCLIP model, which is used to guide an RL agent in achieving success on various specific tasks described in natural language, such as ``collect wood". A similar strategy is used by \citet{Ding2023CLIP4MCAR} which instead relies on a curated dataset of videos to fine-tune the original CLIP model. Beyond MineCraft, VLMs have also been fine-tuned to produce rewards in robotics, by leveraging spatial language descriptions of the visual observations \citep{pmlr-v162-mahmoudieh22a,yang2023learning,du2023video}, by extrapolating annotations from a smaller dataset to a much larger one \citep{Xiao2022RoboticSA} or by computing  cosine similarity at the level of trajectories \citep{Sontakke2023RoboCLIPOD}.  Researchers have also investigated leveraging the CLIP model \citep{alayrac2022flamingo} for computing reward in a zero-shot manner \citep{Cui2022CanFM, rocamonde2023visionlanguage, baumli2023visionlanguage,pmlr-v232-du23b}. 

 The Flamingo \citep{alayrac2022flamingo} model has also been used for its ability to take as input both text and image, and output text. This contrasts with approaches that leverage the latent embeddings of the CLIP model to compute a cosine similarity. Flamingo can indeed answer with coherent textual responses to queries from the user. This ability was leveraged by \citet{pmlr-v232-du23b} to use a fine-tuned Flamingo model as a success detector for embodied robotic tasks. The model is presented with observations from the environment and is asked to infer  whether a task was completed, essentially performing visual question answering. Most of these methods require querying the model multiple times per episodes in order to derive a reward function. This inference cost comes on top of the need to first perform a fine-tuning stage, which may be impractical for larger VLMs. Another important difference with previous work is the fact that most of these methods do not use the VLMs for their (limited) reasoning capabilities, but rather for their embedding space. One exception is \citet{pmlr-v232-du23b}, however their fine-tuned model only outputs simple binary answers. Instead, \namemodel leverages a VLM in a zero-shot manner for crafting reward functions via multiple rounds of prompting through which the VLM can reason about the environment, the subgoals and how to achieve them.  Finally, none of the related methods provide interpretable solutions (as programs or otherwise) for the reward function being generated by the VLM. As \namemodel produces code functions, it provides additional reliability before deployment for inspection whenver required.


\subsection{Code as Policies and Rewards}
Previous work has prompted LLMs to output code whose execution produces robot control actions~\citep{liang2023code}. While this is an interesting approach, policies can be obtained through many other effective techniques, including by running RL algorithms. On the other hand, obtaining reward functions in real-world applications is much more difficult~\citep{dulac2021challenges,10.5555/3491440.3491661, devidze2021explicable, kwon2023reward,ratner2018simplifying}. Approaches similar to code as policies have been briefly explored in LLMs to provide reward in some restricted tasks~\citep{ma2023eureka,kwon2023reward}, but without incorporating image-based observations, which are common in robotics and other applications. To our knowledge, this work is the first to propose the idea of using VLMs to generated code implementing reward functions.


\section{Preliminaries}
In this section we define notation and briefly review relevant reinforcement learning concepts.

Sequential decision making problems are often modeled as Markov Decision Processes (MDPs) \citep{bams/1183519147}. An MDP is a 5-tuple $\langle S,A,T,R,\gamma\rangle$, where $S$ is the state space, $A$ is the action space, $T: S \times A \times S \to [0, 1]$ is the transition function, $R:S \times A \to \mathbb{R}$ is the reward function and $0<\gamma<1$ is the discount factor. At each time step $t$, the agent executes an action $a_t \in A$ in state $s_t \in S$ to get a reward $r_t=R(s_t,a_t) \in R$, and moves to another state $s_{t+1}$ with probability given by $T(s_t, a_t, s_{t+1})$. The objective is to find a policy $\pi: S \to A$ from the space of all possible policies $\Pi$ so as to maximize the expected return (discounted sum of rewards), $\max \displaystyle \mathop{\mathbb{E}}_{\pi \in \Pi} \sum_{t=0}^\infty \gamma^{t} r_t $, where  $a_t = \pi(s_t)$. The reward function $R$ is conceptualized as describing the goal of the agent, and is usually thought of as part of the environment. However, in practice, $R$ usually has to be provided by a domain expert. While many reward functions can share the same optimal policy, having a more informative reward function, which guides the agent at every step, can significantly improve the speed of learning.

Nevertheless, once a good reward function is available, an array of RL algorithms, ranging from model-based~\citep{doya2002multiple} to value-based~\citep{sutton2018reinforcement} and policy-based~\citep{williams1992simple}, can be  applied to find a policy that maximizes the expected future returns.



\begin{figure*}[h]
\centering
    \includegraphics[height=7cm]{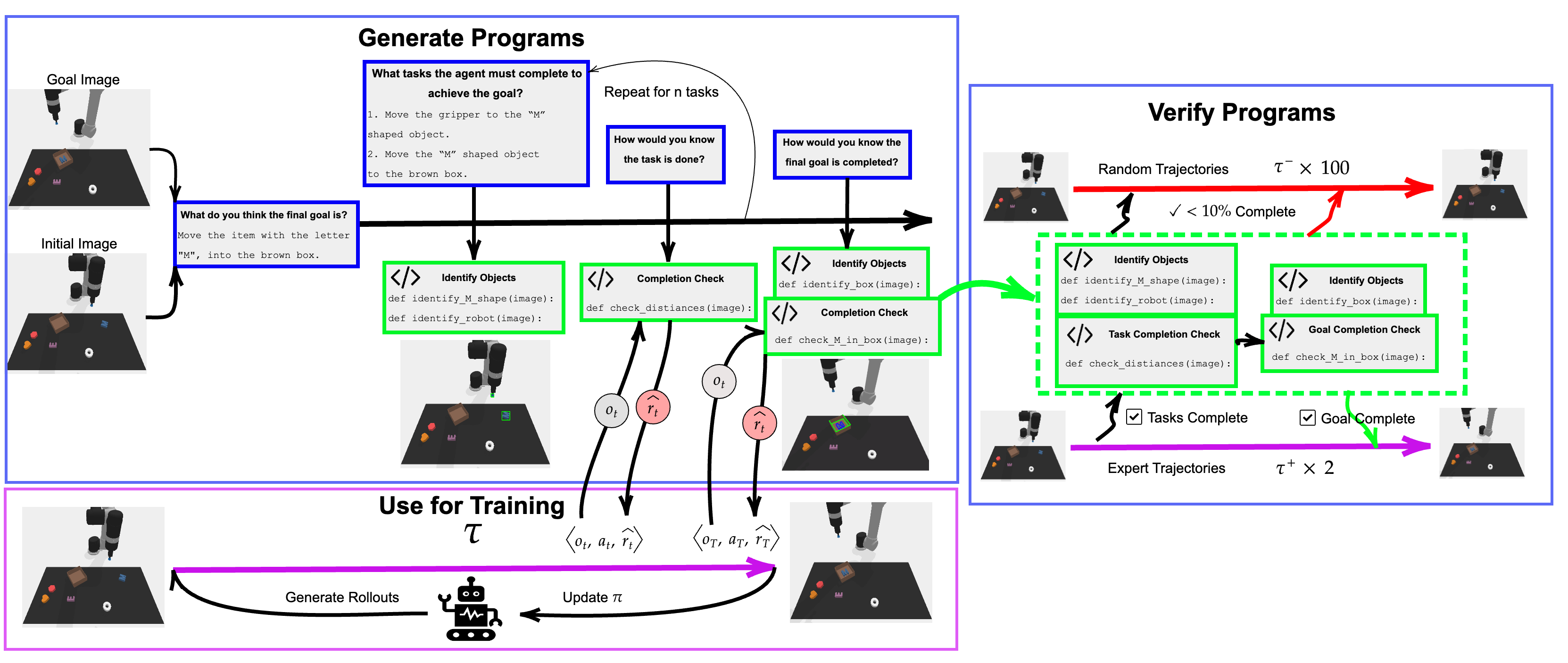}
    {\caption{Complete pipeline of \namemodel, describing how code blocks for sub-tasks and rewards are generated.  The top portion is the reward script generation pipeline, which uses the VLM, and the bottom portion is the RL training loop. The feedback loop is shown on the right and is used to determine if the task and goal code blocks are correct.  The middle portion in green represents the generated scripts from the VLM.  The task completion scripts are applied to random and expert trajectories to compute if the task was completed or not.  All tasks should be completed in expert trajectories and rarely completed in random trajectories.} \label{fig:pipeline}}
\end{figure*} 

A key challenge for RL algorithms is to learn and plan over long horizons, especially when rewards are sparse. The options  framework~\cite{SUTTON1999181} provides a formalism for temporal abstraction, which helps both exploration and credit assignment in such problems. The main idea is to break the full problem that the agent aims to solve into sub-tasks, each of which usually has its own reward function and can be completed by a separate policy. Formally, each option $\omega$ is defined by 3-tuple $\langle \mathcal{I}_{\omega}, \pi_{\omega}, \beta_{\omega}\rangle$, where $\mathcal{I}_{\omega} \subset S$ is  the set of states where $\omega$ is applicable, $\pi_w$ is the policy used when option $\omega$ is executing and $\beta_{\omega}: S \to [0, 1]$ indicates the probability that option $\omega$ terminates in each state  While this framework gives the necessary tools for solving a complex task if the sub-tasks are defined, it remains  difficult to discover options automatically, end-to-end from data \citep{ Bacon_Harb_Precup_2017, Harb2017WhenWI, klissarov2021flexible}.Many works focus instead on using pre-trained option policies \citep{huang2022inner,Ahn2022DoAI}. In this work, we use the generalization capabilities of VLMs to define sub-tasks for solving a complex task, which in turn leads to  dense reward functions for training RL policies later. Our approach can be viewed as using the VLM to generate code corresponding to the termination condition of the option, as well as to the reward function that can be used to learn the policy of the option.

\section{Proposed approach}
 In this section, we present our framework, \namemodel, to autonomously craft options (sub-tasks) and reward functions through code generation using the GPT-4 web interface. Figure \ref{fig:pipeline} presents an overview of our approach. Our framework consists of three distinct stages: a) \textit{Generate Programs}, b) \textit{Verify Programs}, and c) \textit{RL Training}. During the first stage, we use pre-trained VLMs to generate sub-tasks and reward definitions as programs. 
  During the second stage, we verify and refine the correctness of the generated sub-tasks and reward functions by evaluating a handful of expert trajectories and random trajectories. During the this stage, we use the reward function to train RL agents. We explain the details of each stage in the subsequent sections


\subsection{Generating Rewards and Sub-tasks }
We primarily rely on sequential prompting for generating programs that represent sub-tasks and reward functions. Although VLMs have good understanding of visual inputs, we have found that they are not yet reliable enough to generate correct rewards in a zero-shot manner for complex tasks, like those under consideration in RL benchmarks. To handle this, we prompt the pre-trained VLMs in a sequential manner, incrementally building up the necessary information to specify the reward functions, while using verification checks wherever applicable. 

Concretely, the sequence of prompts obey the following structure across all tasks presented in this work. Given only the initial frame of a trajectory, we first prompt the VLM to identify what part represents the agent in the image and ask it to provide a script returning the agent's location.  We then provide the VLM with an image where the agent has reached its goal state and ask the VLM to infer what the task was.  Next, we ask the VLM to describe all the relevant objects in the image.  Given this sequential build-up of relevant information and objects, we ask the VLM to provide $n$ sequential sub-tasks (with $n$ being a tune-able parameter), that are actionable by the agent and will result in the final goal completion.  An example set of tasks is shown in Figure \ref{fig:pipeline}.  For each sub-task, we prompt the VLM to infer the subset of objects that are relevant for verifying the completion of this particular sub-task, followed by generating a program for identifying the relevant objects. We finally ask the VLM to provide a program for identifying if the sub-task is complete. We verify this program using the verification pipeline in Subsection \ref{verpipeline}. This serves as an initial set of functions that can be thought of as implementing both initiation and  termination conditions for options corresponding to each sub-task. In other words, the objects in the image, identified by the VLM, determine both which options are affordable~\citep{Khetarpal2020OptionsOI, Ahn2022DoAI} as well as when options terminate.  For reaching the final goal, we utilize the same approach to determine the relevant objects and a program to identify if the final goal is completed. 
We show the full prompting pipeline in Appendix A and B as well as an example interaction with the VLM in Appendix C.

\subsection{Verification using Expert and Random trajectories}
\label{verpipeline}

A possibility for obtaining reliable programs is to ask humans to visually inspect the correctness of the code through a set of observations from the environments. While human expertise can be invaluable, it can limit the scaling of this approach, as generated scripts are complex, especially in real world environments. Therefore, we automate the verification of generated scripts through a verification pipeline.

To automatically verify the correctness of the generated programs described in the previous section, we use 2 trajectories generated by an expert agent and $100$ randomly generated trajectories.  We execute the generated object identification and sub-task programs  on the random and expert trajectories.  Every sub-task must be completed in the expert trajectories and completed in less than $p\%$ of the random trajectories (where $p$ is a tune-able parameter). It is important to note that it is very important to use random trajectories as negative data, in order to catch any false positive sub-tasks generated by the model. This verification stage is shown on the right in Figure \ref{fig:pipeline}.  If the generated programs do not pass this verification stage, the VLM is prompted anew and asked to refine the program for sub-tasks which did not pass the automated  verification test. This allows us to ensure the correctness of the sub-tasks and reward functions.

\subsection{Using Generated Programs in the RL loop}
Once the generated programs are verified, we use them to improve the learning efficiency of an RL agent. The programs generated for sub-task completion are added as auxiliary reward functions in addition to the original environment reward. Specifically, once the proposed sub-task is determined to be completed, an additional auxiliary reward ($r_{aux}$) is provided to the agent and we move to the determining if the next task is completed or not. Once all sub-tasks are completed, we call the final goal completion script. The assembly of all of the VLM's proposed sub-goals scripts leads to a dense reward model. These sub-task completion checks could also be thought of as option termination conditions. In that view, the VLM generates the option initiation and option termination conditions as programs, while the option policy is learned with reinforcement learning. It should be noted that we can also leverage hierarchical learning frameworks like options with the generated programs but in this work, we stick to auxiliary rewards for simplicity.

\section{Experiments}
We conduct experiments with the \namemodel-generated reward function to show two major improvements: (1) \namemodel can transform a sparse reward function into a set of dense reward functions for each sub-task.  These per-task rewards are much more efficient for training RL agents than the environment provided sparse rewards.  We show these results in discrete action grid environments and in robotic control tasks. (2) \namemodel can generate a reward function for difficult high dimensional robotic environments only from the an image of the initial and completion state.

\subsection{Experimental Procedure}
We use the GPT-4 web interface for all experiments, which allows access to VLM capabilities and the uploading of images.  This product is available with a subscription of only \$20 USD per month.  We employ the mentioned set of prompts and pipeline to generate scripts which discover tasks and goals, identify the relevant objects in each task and goal and check its completion. Note that this design is highly cost efficient and can be handled solely from the GPT-4 web interface without requiring additional cost. 

\begin{table*}[h!]
\caption{This table shows the number of failed attempts (Failed) that the VLM had for each task in each environment.  We also list the number of response regenerations (Regens) required due to the internal timeout for generating a response.  Failed attempts could be due to the automated verification pipeline where we proposed, or the first-pass check with the VLM not identifying any object in the initial image (returning False during object identification).  N/R indicates that this object was not used in any task or goal completion scripts.  We also provide a brief textual description of the strategy employed in the generated identification and completion check script.  For example vertices means the shape was identified by checking the number of vertices (4 is typically a rectangle or square, $>4$ is a more complex shape). $n$ denotes the number of tasks asked for the in the pipeline. }
\label{sample-table}
\vskip 0.15in
\begin{center}
\begin{small}
\begin{sc}
\begin{tabular}{lcccccr}
\toprule
Environment & Task & Failed & Regens & Strategy Employed & n\\
\midrule
MiniGrid    & Identify Agent &  $2$ & $ 0$ & Vertices/Colour & 3 \\
 & Identify Key & $2$ & 0 & Vertices/Colour/Size\\
    & Check Key Complete& $2$ & $0$ & Key Disappeared\\
    & Identify Door &  $0$ & $0$  & Vertices/Colour      \\
     &  Check Door Complete & $4$ & $0$ & Door Shape Change\\
      & Identify Goal & $0$& $0$ & Vertices/Colour \\
      & Check Goal Completion & 3 &   0  & Agent Proximity    \\
PandasGym   &  Identify Agent & N/R & N/R & Vertices/Colour/Size & 2\\
        &  Identify Green Block & 3 & 2 & Colour/Size \\
        &  Check Green Block Moved & 3 & 1 & Moved from Initial Pos \\
        &  Identify Yellow Block & 1 & 0 & Colour/Size \\
        &  Check Goal Completion & 2 & 1 & Contour Overlap \\
separating-piles   &  Identify Yellow Square & 2 & 0 & Colour/Size & 1 \\
   &  Identify Blue Blocks & 3 & 1 & Colour \\
   &  Check Goal Completion & 5 & 4 &  Contour overlap\\
put-blocks-in-bowls  &  Identify Green Bowls & 3 & 0 &  Colour/Size/Circularity & 1 \\
  &  Identify Red Blocks & 1 & 2 & Colour/Size\\
  &  Check Goal Completion & 6 & 2 &Contour overlap \\
packing-shapes  &  Identify M Shaped Object & 11 & 0 & Vertices/Colour & 1\\
  &  Identify Brown Box & 2 & 1 & Colour/Size\\
  &  Check Goal Completion & 4 & 1 &Contour Overlap \\
\bottomrule
\label{tab}
\end{tabular}
\end{sc}
\end{small}
\end{center}
\vskip -0.3in
\end{table*}

Since the VLMs are not accurate enough to generate reward and sub-task program in first attempt, we need to run the generation pipeline multiple times till our verification pipeline succeeds. In Table \ref{tab}, we list the number of failed attempts for each suggested task and object identification. We also list the number of regenerations required due to the internal GPT-4 timeout for taking too long to complete a response. It also shows that VLMs are good enough to generate these programs in a few attempts and efficacy of our pipeline to catch false positives. We show a few inaccurate programs caught by our verification pipeline in Appendix D as well as usefulness of random trajectories.

\subsection{Manual Steps}
The VLM gives us a set of functions, usually separate blocks of code to identify the objects and check task completion.  We do not ask the VLM to assemble everything automatically into a single class. In the current implementation, we only assemble these scripts together into a single class which has a callable reward function  (as one of its method). We use the generated sub-task functions by VLMs to create a very simple reward function which calls the VLM generated functions (for sub-tasks) and we give a reward of $1 / n$ for each of the $n$ tasks.  This step also requires us to store a boolean variable for each trajectory indicating if the task is done, so we can move onto checking the next task. These variables are reset at the start of every episode. While this process of having human-in-the-loop for assembling sub-task scripts can be seen as a potential limitation for scaling to more complex tasks, we also believe this can be addressed in future work by prompting the VLM to assemble the final script or using the sub-task completion programs within the option framework where the learning algorithm can directly use these programs for option termination. It is not achieved in this work due to practical limitation of timeouts in the web interface of GPT-4 for large contexts and programs. 

We believe this work realises the potential of using VLMs to generate programs as a first step. Building completely automated end-to-end pipeline without any human in the loop is an interesting direction for future research. As VLMs continue to scale and with a proper API access to the VLM, we believe that we could eliminate this manual task.


\subsection{MiniGrid}
\begin{figure*}[h!]%
\begin{center}
    \subfigure{\includegraphics[height=3.4cm]{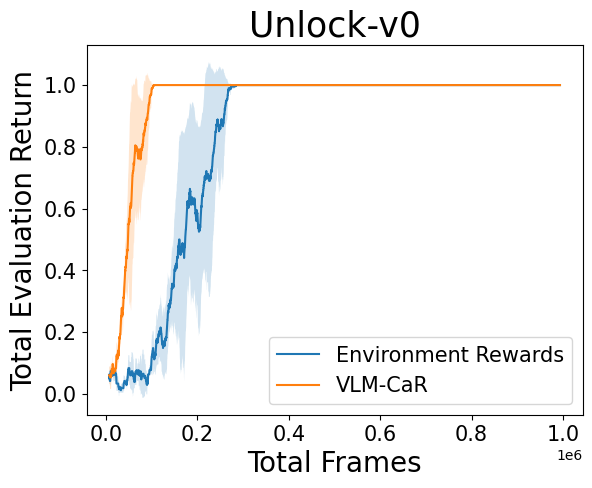}} \hspace{0.0cm}%
        \subfigure{\includegraphics[height=3.4cm]{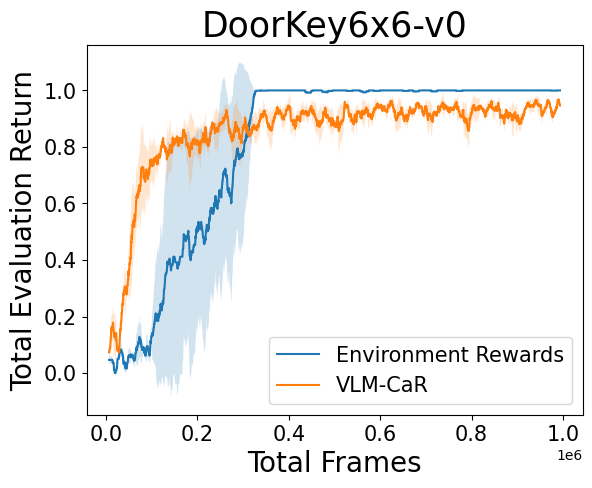}} \hspace{0.0cm}%
        \subfigure{\includegraphics[height=3.4cm]{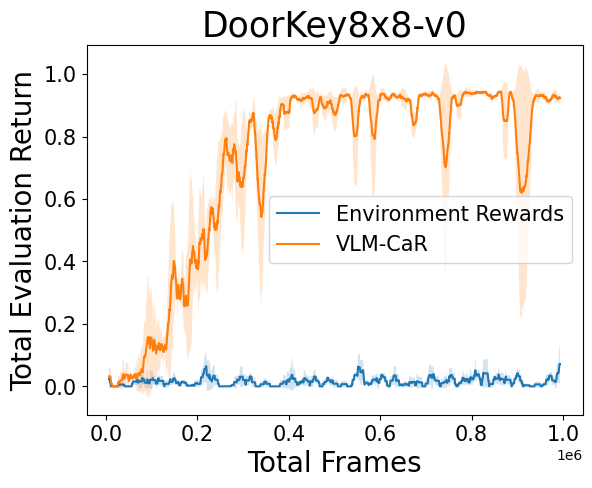}} \hspace{0.0cm}%
    \subfigure{\includegraphics[height=3.4cm]{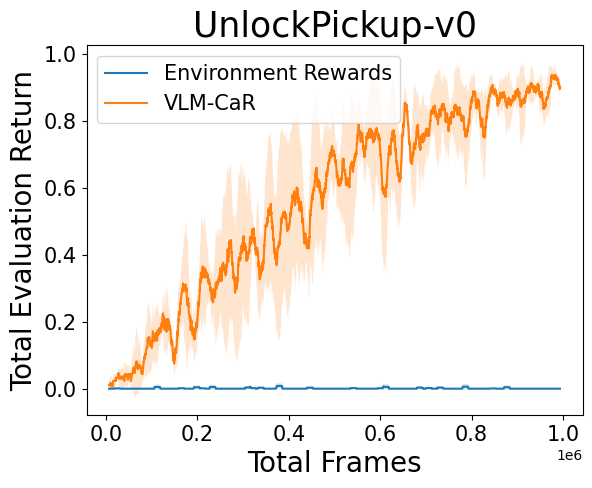}} \hspace{0.0cm}%

\hspace{0.0cm}%
        \end{center}
        \vspace{-8mm}
        \caption{The online episodic mean reward evaluated over 5 episodes every 250 steps for MiniGrid RL tasks.  We show the average over $3$ random seeds.  $1M$ environment step interactions are used. The shaded area shows the standard error. Agents trained using rewards generated by \namemodel perform better than the sparse environment reward.  In some tasks, sparse rewards are not sufficient for any meaningful performance whereas \namemodel rewards allow the agent to solve the task.}
        \label{minigrid}
        \vspace{-2mm}
\end{figure*}

We first show the Gym-MiniGrid \citep{MinigridMiniworld23} set of partially-observable environments, for which rewards are sparse and there is a non-zero reward only when the agent completes the task. The agent is only given a local ego-centric view of its environment. For example, in DoorKey a key must first be obtained and then a door needs to be opened leading to another room with the goal state. We use the sparse environment based reward function for evaluation in our experiments.  In all MiniGrid experiments we use proximal policy optimization (PPO) \citep{schulman2017proximal} to train our RL agent. In this environment, VLM is able to generate sub-tasks such as as obtaining key, reaching door and then reaching final goal. Each sub-task is assigned an auxiliary reward of $1 / 2$ with a final reward of $1$ for completion of the goal. We show that the \namemodel generated auxilary rewards are useful and accelerate learning in contrast to just using the sparse reward of environment in Figure \ref{minigrid}.  The agent trained with \namemodel rewards is able to complete the difficult DoorKey8x8 environment whereas traditional PPO is not.  In all environments we show remarkable improvements in training efficiency. 

\textbf{Reward Strategies: } For DoorKey and Unlock, the agent is identified by looking for a red triangle, with $3$ vertices in the contour.  The Key is identified by searching for a complex yellow shaped object, identified by an object that has greater than $4$ vertices.  The door is identified as searching for a yellow rectangle by determining color and the number of vertices. Completion checks consist of identifying if the key has disappeared from the image to suggest completion of the first task and if the door has opened to check completion of the 2nd task.  The ``is\_door\_open" script is remarkably intuitive, checking if the door has changed shape from an approximate square to a smaller rectangle. We provide all generated scripts in the Supplementary Material.

\begin{figure*}[h!]%
\begin{center}
    \subfigure{\includegraphics[height=4cm]{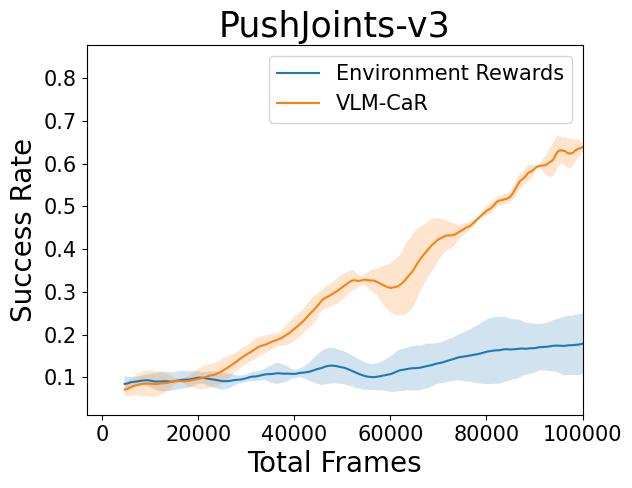}} \hspace{0.0cm}%
        \hspace{0.0cm}%
    \subfigure{\includegraphics[height=4cm]{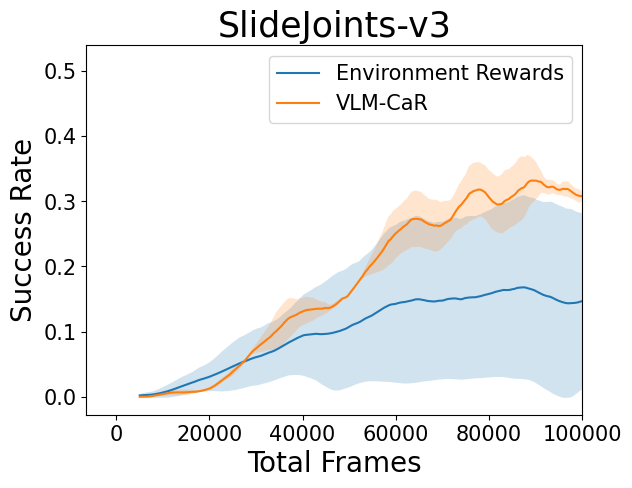}} \hspace{0.0cm}%

        \end{center}
        \vspace{-6mm}
        \caption{The success rate in completing the final task in Pandas-Gym environments. $5$ random seeds are shown. The shaded area is the standard deviation. RL agent trained on dense reward generated by \namemodel generally performs better than RL agent trained on sparse environment rewards.}
        \label{pandas}
        \vspace{-2mm}
\end{figure*} 

\begin{figure*}[]%
\begin{center}
    \subfigure{\includegraphics[height=4cm]{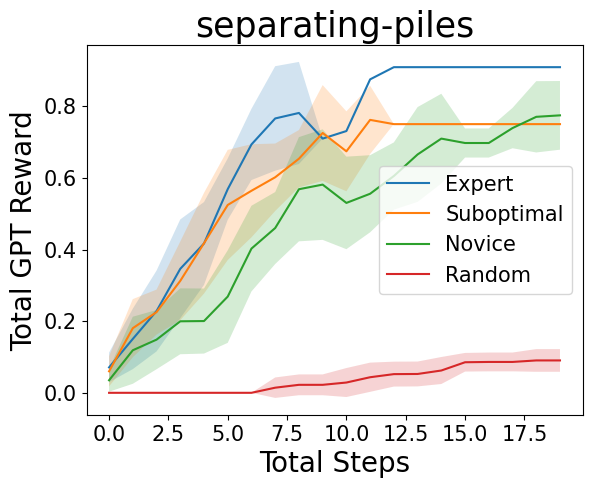}} \hspace{0.0cm}%
        \hspace{0.0cm}%
    \subfigure{\includegraphics[height=4cm]{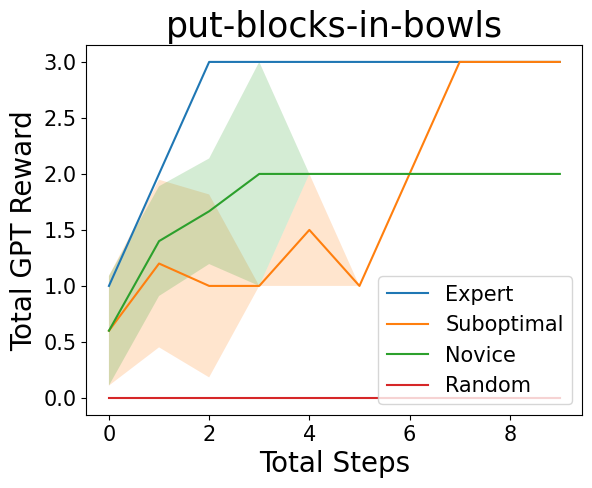}} \hspace{0.0cm}%
    \subfigure{\includegraphics[height=4cm]{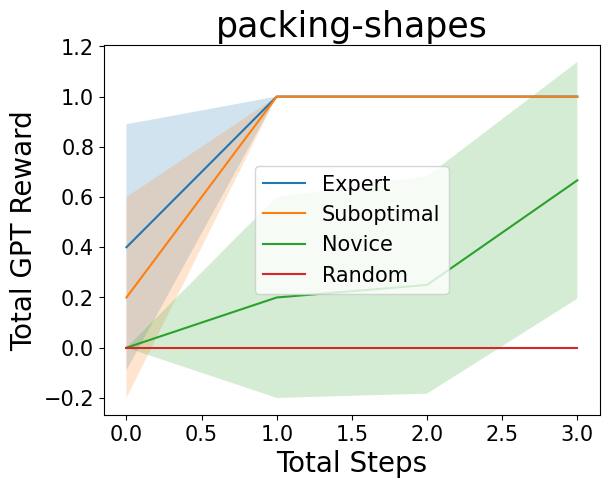}} \hspace{0.0cm}%

        \end{center}
        \vspace{-6mm}
        \caption{The inferred reward using \namemodel for 4 policies (random, novice, sub-optimal, expert) on different CLIPort environments.  CLIPort is trained using imitation learning so we evaluate or reward function on 4 policies of varying skill levels. An action is taken randomly 30\% of the time in the suboptimal policy and 50\% of the time in the novice policy. The reward inferred by \namemodel well reflects the training process and performance of different policies.}
        \label{cliport}
        \vspace{-2mm}
\end{figure*}

\subsection{Pandas-Gym}
Pandas-Gym provides a simulation environment to benchmark RL agents on a variety of continuous control tasks \citep{gallouedec2021pandagym}.  In these experiments, we control the joints of the robot.  We examine two tasks in these experiments, in each task the agent must move a green cubic object into a yellow target area either by pushing (PandasPush) or sliding (PandasSlide) it across a long surface.  We show results in Figure \ref{pandas} where the \namemodel rewards improve the success rate in contrast to just using environmental rewards. It also improves samply efficiency for a given success rate.  We use Truncated Quantile Critics (TQC) \citep{10.5555/3524938.3525453} to train our policies, the current optimal benchmark for Pandas-Gym.

\textbf{Reward Strategies}:  In the first push task, the green and yellow blocks are identified correctly.  The first task is to make contact with the green cube.  The completion check proposed is to verify if the green block has moved by comparing its current position to its position in the initial image.  The final reward function for the goal proposed by the VLM is to determine the Euclidean distance between goal and block relative to initial image.  It is then computed at each step afterwards, determining the progress that the agent has made in moving the block to the goal.  The reward function is $r = (d_{\text{initial}} - d_{\text{current}}) / d_{\text{initial}}$, where $d$ indicates the Euclidean distance.  Since objects are of the same colour and shape in all tasks, this script was found to be generalizable to both PandasPush and PandasSlide.

\subsection{CLIPort}

Lastly, we focus on robotic environments utilized in CLIPort \citep{shridhar2021cliport}. These are simulated robotic tasks and utilize a Universal Robot (UR5e) with a suction gripper. CLIPort agents are trained using imitation learning, so we cannot access the online training procedure like we illustrated in previous two environments. In these experiments, we illustrate that \namemodel can correctly generate an effective reward function for this complex environment. Specifically, we evaluate 4 agents from the pre-trained checkpoints of varying skill levels i.e random, novice, sub-optimal and expert. We roll-out trajectories from these different policies and evaluate our \namemodel generated reward function on these trajectories.

For robotic tasks, we ask the VLM to give an incremental reward model if it determines that is more appropriate. For the Robotic Goal Completion Pipeline, we only provide a reward function for completing the final goal.  This reward function is usually dense in many robotic tasks, where for example the agent must move multiple objects to a target location.  We observe that it is common for the VLM to propose the Euclidean distance of the targets from the goal location as the goal reward model in this case.  

The setup provides a complex environment for evaluating the ability to ground natural language concepts like colors and object categories. The input is a top-down RGB-D image from 3 cameras positioned around a rectangular table: one in the front, one on the left, and one on the right, all pointing towards the center of the table.  For \namemodel, we use the camera overlooking the center of the table from a 45 degree angle to provide images of the initial and goal state.  We examine 3 tasks in this UR5e environment.  The first, \url{separating-piles}, the agent must sort blue blocks into a yellow square shaped region.    In \url{put-blocks-in-bowls}, the agent must move $2$ or $3$ red blocks into the green bowls.  In \url{packing-shapes}, the agent must move a blue "M" shaped object into a large brown box.  For simplicity, we fix the colours and shapes of the relevant objects in each task, since a component of our VLM generated reward function utilizes the identification of colours as well as shapes.  CLIPort is trained utilizing textual descriptions of the task, for example ``sort the red blocks into the green square".  For utilizing different coloured objects, it is easy to see that we could modify \namemodel to correctly identify different coloured objects using the textual descriptions.

\textbf{Reward Strategies:} For \url{separating-piles}, \namemodel generates a reward function that first identifies the yellow square from the initial frame and saves its contour for future use.  \namemodel then identifies all the blue block contours in the image and give a numerical reward that is the fraction of blue blocks inside the yellow square to the portion outside.  For \url{put-blocks-in-bowls}, \namemodel first identifies the red cubic shaped objects.  The green circular shaped bowls are then identified and the contours are stored from the initial image.  \namemodel then gives a reward of $1$ for each red block that is identified in the green bowl contour.  For \url{packing-shapes}, \namemodel first identifies the blue "M" shaped object by checking if the shape is more complex than a rectangle (with vertices $> 4$) and identifying it by it's blue color and a relatively large size.  \namemodel then identifies the brown box by size, if it is rectangular and by color.  The reward is $1$ for identifying the "M" shaped object inside a bounding contour of the box.  We utilize the robotic-goal-completion pipeline for these experiments. Our experiments in Figure \ref{cliport} indicate that \namemodel generated reward function is able to separate different policies on these different tasks. All the policies are clearly separable in separating-piles task. For tasks of put-blocks-bowls and packing-shapes, sub-optimal and expert policies are close as per \namemodel reward function but both are clearly separated from novice and random policies showing the effectiveness of reward functions.

\section{Limitations}
Throughout the utilization of our pipelines we found numerous limitations as well as some very interesting behaviours that suggest future strategies. Fundamentally, the image processing approaches suggested by the VLMs focus mainly on identification by colour and shape.  Identifying objects that have very similar shapes and colours is extremely difficult given current models. For example, in MetaWorld \citep{yu2019meta} the robot is nearly an identical color of the objects it interacts with making the problem far more challenging.  We find that color detection is sometimes necessary, even with complex shape detection procedures.  Resolution also impacts detection algorithms.  This work is not a suggestion that our method will work on every environment, only a suggestion that this is a successful direction for many types of RL tasks.

Shadows as well as the overlapping of objects, also create issues in shape, color and edge detection algorithms.  We therefore prompt our VLM to store the locations of expected static objects from the initial image and to try and approximate shapes in contours as they may not be exact (or incomplete). We can do this by finding a convex hull of every contour which would always be closed and then approximating it down.  In more complex tasks, we may have to show the VLM a full expert trajectory instead of the final and initial images to infer the tasks and rewards.

Our approach also requires manual assembly of the VLM provided scripts to create the final callable program, which we hope to eliminate as VLMs continue to scale. We hope as VLMs become more accurate in the future, these challenges will become easy to overcome.  We also note that with access to a VLM with unlimited interactions and many iterations of our verification pipeline, it is likely that we could develop more complex strategies easier. 

Despite these limitations, we see our method as an important step towards solving real-world tasks with natural image-based observations using reinforcement learning, empowered by \namemodel's code as reward capability.

\section{Future Work and Conclusion}

\namemodel is an intuitive approach to reward design in an easily accessible and tune-able manner.  We showcase both automated task generation, automated reward function design and interpret-able reward functions in a wide variety of complex tasks.  These rewards are found to be superior to environmental rewards in sparse reward tasks.

During our experiments, we found the VLM to suggest implementing a pathfinding algorithm such as A* for finding a path to the goal state or key in MiniGrid.  In one instance the VLM listed the correct sequence of steps given only the initial and goal image. We believe that our methods could also be used to generate progress models ~\citep{bruce&al23progress, mazoure&al23} or generate affordances ~\citep{khetarpal&al20} as programs in future.

\section{Impact Statement}
The goal of the work presented in the paper is to advance the field of RL. There are many potential societal consequences of RL, but specific impacts are therefore difficult to highlight here.

\section*{Acknowledgements}
We would like to thank Gheorghe Comanici for reviewing the draft of this work and providing critical feedback which helped improve the quality of manuscript significantly.

\nocite{langley00}

\bibliography{example_paper}
\bibliographystyle{icml2024}

\newpage
\appendix
\onecolumn

\section{Task Completion Prompting Pipeline}
\begin{itemize}
\item In this image, there is an agent that can move throughout the environment.  There may also be other relevant objects that the agent can interact with.  The agent is a \{agent description\}. 

\item (For robotic environments only:) What is the part of the agent that would most likely interact with objects? Give me only one object.

\item Can you write a script to identify the agent object from the image?  You should use the shape, edges, and color of the object to identify it as its possible other objects in the image may have the same color.  This should return the location of the agent along with \{True, False\} if it is found.  This script should not require input from me.
\end{itemize}
(Check verification pipeline)

\textbf{Fail Response: }  Please try again and refine you approach. Please remember to identify objects using edges, shape and colour.  Please examine the shape and colour of this object from the image again.

\textbf{Fail Response (5 or more times): }  Please try again and refine you approach. Please try to simplify your approach by only checking color or simple shapes.

\begin{itemize}

\item This is the correct script to identify the agent.  Please remember this script.  I will refer to it as the "agent\_ID\_script".

\item Can you give a list of the most important objects in this image?  Do not give general objects like the walls or the grid.  Give me a list of objects and concise names with description.

\item I am now going to show you an image of the game when the final goal was completed. 

\item What do you think the final goal is?  Give me only one goal.

\item Now, from this image, can you infer $n$ sequential tasks that the agent must do before it reaches the final goal.  Give me the list of $n$ tasks with descriptions of each task.  This list should be concise and should not contain general behaviors like: navigate the maze, or avoid walls.  The items you have identified in the image may help you come up with this list of tasks.   The tasks should be actionable and concise and must complete the final goal.
\end{itemize}
\textbf{Task completion check procedure:}
\begin{itemize}
\item Now for Task 1, what is the most relevant object or objects in this task to reach or interact with.  This item should not be the agent. 

\item Can you write a script to identify this object(s) from the image?  You should use the shape, edges and color of the object to identify it as its possible other objects in the image may have the same color. You can only use the first image I gave you as input to this script.  This script should return the location of the object(s) and {True, False} if it is found.  Test by verifying the number of instances of the object found is correct. This script should not require input from me.
\end{itemize}
(Task completion pipeline)

\textbf{Fail Response: }  Please try again and refine you approach. Please remember to identify objects using edges, shape and colour.  Please examine the shape and colour of this object from the image again.

\begin{itemize}
\item This is the correct script to identify the Task 1 item.  Please remember this script.  I will refer to it as the "task\_1\_o1\_ID\_script".  

\item How would you know if Task 1 is done?  Please propose one best guess of the check for completion.  The type of check for completion you use should be implementable using a python script.  Please describe this check. The script can only use the first image I gave you.  You are allowed to compare this to the image from the initial frame.

\item Are there any other objects that are absolutely essential to interact with or must be identified to check if this task has been completed.  Only list absolutely essential objects.
\end{itemize}
(Identify objects using the object identification technique and label these object scripts accordingly)
\begin{itemize}
\item Please implement this technique, you will only have access to a single frame at a time. If you would like to access the initial state and compare it to the current state or store information from the initial state, this is allowed. Please use the same “task\_1\_o1\_ID\_script”, \dots as part of your implementation.  The “agent\_ID\_script” may also help you. You can only use the first image I gave you as input to this script.  This script must return \{True, False\}. It may be useful to store the location of expected static objects from the initial image. This script should not require input from me.  
\end{itemize}

(Repeat task completion check procedure for $n$ tasks.)

\textbf{Goal completion check procedure:}
\begin{itemize}

\item How would you know if the goal is done?  Please propose one best guess of the check for completion.  The type of check for completion you use should be implementable using a python script.  Please describe this check. The script can only use the first image I gave you.  You are allowed to compare this to the image from the initial frame.  Can you please tell me first if you must identify any new objects for this task.

\item Are there any objects that you have not identified so far that must be identified to check the completion of the goal.
\end{itemize}
(Identify objects using the object identification technique and label these object scripts accordingly)
\begin{itemize}
\item Please implement this technique, you will only have access to a single frame at a time. If you would like to access the initial state and compare it to the current state or store information from the initial state, this is allowed. Please use the same “task\_1\_o1\_ID\_script” or, \dots \{all other object ID scripts\} \dots as part of your implementation.  The “agent\_ID\_script” may also help you. You can only use the first image I gave you as input to this script.  This script must return \{True, False\}. It may be useful to store the location of expected static objects from the initial image.  I will show you two images, the first image is of the initial state and the 2nd image is after the goal completion.  You must return False on the first image and true on the 2nd. A few useful techniques: Use the contours to determine if an object is inside of another, don’t approximate rectangles or radii.  Try to approximate shapes in contours as they may not be exact (or incomplete). You can do this by finding a convex hull of every contour which would always be closed and then approximating it down. Checking the shape of objects is useful once you have identified them by color.  This script should not require input from me. 

\item For robotic tasks: Could you also propose a real valued reward function that is incremental if possible?
\end{itemize}

\textbf{Fail Response: }  Please try again and refine you approach.

\section{Robotic Goal Prompting Pipeline}

\begin{itemize}
\item In this image, there is an agent that can move throughout the environment.  there may also be other relevant objects that the agent can interact with.  The agent is a robotic arm. 

\item What is the part of the agent that would most likely interact with objects? Give me only one object.

\item Can you give a list of the most important objects in this image?  Do not give general objects like the walls or the grid.  Give me a list of objects and concise names with description.

\item I am now going to show you an image of the game when the final goal was completed.

\item What do you think the final goal is?  Give me only one goal.

\item How would you know if the goal is done?  Please propose one best guess of the check for completion.  The type of check for completion you use should be implementable using a python script.  Please describe this check. The script can only use the first image I gave you.  You are allowed to compare this to the image from the initial frame.  

\item Give me the list of essential objects that must be identified to know if the goal has been completed.  Do not include the agent in this list.  Only give objects that are absolutely essential.
\end{itemize}
(Identify objects using the object identification technique and label these object scripts accordingly)
\begin{itemize}
\item Are there any objects that you have not identified so far that must be identified to check the completion of the goal.

\end{itemize}
(Identify objects using the object identification technique and label these object scripts accordingly)
\begin{itemize}
 \item How would you know if the goal is done?  Please propose one best guess of the check for completion.  The type of check for completion you use should be implementable using a python script.  Please describe this check. The script can only use the first image I gave you.  You are allowed to compare this to the image from the initial frame.  You can store the location of objects expected to be stationary.

\item Please implement this technique, you will only have access to a single frame at a time. If you would like to access the initial state and compare it to the current state or store information from the initial state, this is allowed. Please use the same “task\_1\_o1\_ID\_script” or, \dots \{all other object ID scripts\} \dots as part of your implementation. You can only use the first image I gave you as input to this script.  This script must return \{True, False\}. It may be useful to store the location of expected static objects from the initial image.  I will show you two images, the first image is of the initial state and the 2nd image is after the goal completion.  You must return False on the first image and true on the 2nd. A few useful techniques: Use the contours to determine if an object is inside of another, don’t approximate rectangles or radii.  Try to approximate shapes in contours as they may not be exact (or incomplete). You can do this by finding a convex hull of every contour which would always be closed and then approximating it down. Checking the shape of objects is useful once you have identified them by color. This script should not require input from me.  

\item Could you also propose a real valued reward function that is incremental if possible?

\end{itemize}

\textbf{Fail Response: }  Please try again and refine you approach.  

\section{Response Examples}
Only successful interactions are shown here.  The approach may have required more iterations and fail responses.  

We show the CLIPort put-blocks-in-bowls task here.

\begin{itemize}
    \item \user{User:} In this image, there is an agent that can move throughout the environment.  there may also be other relevant objects that the agent can interact with.  The agent is a robotic arm. 
    \item \gpt{VLM:} The image depicts a robotic arm positioned above a flat surface with various colored objects. There are green cylindrical containers, each with a different colored cube inside, except for one which appears to be empty. There are also red cubes outside the containers, suggesting that the task for the robot could be to sort these cubes into their matching colored containers. The robotic arm itself is equipped with a gripping tool that it can use to pick up and move the objects. This setup is typical in robotics and artificial intelligence research, where robots are trained to perform sorting tasks which can serve as benchmarks for their dexterity and decision-making capabilities.
    \item \user{User:} What is the part of the agent that would most likely interact with objects? Give me only one object.
    \item \gpt{VLM:} The part of the agent that would most likely interact with objects is the gripper at the end of the robotic arm.
    \item \user{User:} Can you give a list of the most important objects in this image?  Do not give general objects like the walls or the grid.  Give me a list of objects and concise names with description.
    \item 
\gpt{VLM:} Certainly! Here's a list of the most important objects in the image:
\begin{enumerate}
\item Robotic Arm: A multi-jointed mechanical arm with a gripper, positioned over the surface.
\item Green Containers: Cylindrical objects with open tops, designed to hold cubes.
\item Red Cubes: Solid red-colored blocks, some placed inside the containers, some on the surface.
\item Blue Cube: A single solid blue-colored block, placed inside one of the green containers.
\item Orange Cube: A single solid orange-colored block, placed inside one of the green containers.
\item Yellow Cube: A single solid yellow-colored block, placed inside one of the green containers.
\item Purple Cube: A single solid purple-colored block, placed inside one of the green containers.
\item Empty Green Container: One of the cylindrical green containers without a cube inside it.
\end{enumerate}
\item \user{User:} I am now going to show you an image of the game when the final goal was completed.
\item \gpt{VLM}: The final goal appears to be placing each colored cube into the matching colored container. Based on the setup, the specific goal completed would be: Place the red cubes into the green containers. 
\item \user{User:}  How would you know if the goal is done?  Please propose one best guess of the check for completion.  The type of check for completion you use should be implementable using a python script.  Please describe this check. The script can only use the first image I gave you.  You are allowed to compare this to the image from the initial frame.
\item \gpt{VLM:} To check if the goal is completed using a Python script, you would implement an image processing solution to compare the initial state of the game (as depicted in the first image) with the final state (as shown in the second image). Here's a proposed method for the check:

Color Detection and Position Mapping: Use a color detection algorithm to identify the colors of the cubes and the containers. You would map the position of each cube relative to the containers.

State Comparison:

Initial State: Detect all cubes and their initial positions outside of the containers.
Final State: Detect all cubes and confirm their positions inside the containers.

Check for Completion:

Create a mapping of cube color to container color.
Verify that each cube is inside a container with a matching color.
Here is a high-level description of what the Python script would do:
\begin{enumerate}
\item Load the images: Use a library like OpenCV to load the initial and final state images.
\item Preprocess images: Convert the images to a format suitable for color detection (e.g., HSV color space).
\item Detect and label objects: Use color range thresholds to detect cubes and containers, and label them based on color.
\item Determine cube positions: Calculate the centroid of each cube and container to determine their positions.
\item Check for red cube placement: Specifically for the red cubes, check if their centroids in the final image lie within the boundaries of the green containers.
\item Result: If all red cubes are within the green containers, the check returns True, indicating the goal is achieved.
Implementing this in a script would involve quite a bit of detail, especially in terms of handling variations in lighting and perspective, but the above method provides a strong basis for such a script.
\end{enumerate}
\item \user{User: }  Give me the list of essential objects that must be identified to know if the goal has been completed.  Do not include the agent in this list.  Only give objects that are absolutely essential.
\item \gpt{VLM:} To determine if the goal has been completed, the essential objects that must be identified are:

Red Cubes: To check if all the red cubes are placed inside the green containers.
Green Containers: To confirm that the red cubes are within their bounds.
\item \user{User: } Can you write a script to identify this object(s) "Number 1" from the image?  You should use the shape, edges and color of the object to identify it as its possible other objects in the image may have the same color. You can only use the first image I gave you as input to this script.  This script should return the location of the object(s) and {True, False} if it is found.  Test by verifying the number of instances of the object found is correct. This script should not require input from me.

\item \gpt{VLM:} The revised script detected two red cubes in the image, and their locations are highlighted with green rectangles. The detection may still not be perfect, but this result shows where the algorithm is currently identifying red cubes. If there's still a discrepancy, further refinement would be needed.

\item \user{User:} This is the correct script to identify the item.  Please remember this script. I will refer to it as "o1\_ID\_script".
\item \gpt{VLM:} Understood. The script used in the last execution to identify red cubes in the image will be referred to as "o1\_ID\_script" for future reference. If you need to use or refer back to this script, just mention its name, and I will know to use the corresponding code.
\item \user{User:} Can you write a script to identify this object(s) "Number 2" from the image?  You should use the shape, edges and color of the object to identify it as its possible other objects in the image may have the same color. You can only use the first image I gave you as input to this script.  This script should return the location of the object(s) and {True, False} if it is found.  Test by verifying the number of instances of the object found is correct. This script should not require input from me.
\item \gpt{VLM:} The refined script has successfully identified two objects that could be the green containers in the image. Here are their locations:

Green Container 1: Located at (x: 110, y: 356), with a width of 69 pixels and a height of 64 pixels.
Green Container 2: Located at (x: 325, y: 290), with a width of 59 pixels and a height of 52 pixels.
The script found two instances, which likely correspond to the green containers on the table. 

\item \user{User: } This is the correct script to identify the item.  Please remember this script. I will refer to it as "o2\_ID\_script".

\item \gpt{VLM:} I will remember the script used to identify the green containers as "o2\_ID\_script". If you need to refer back to it or use it in the future, just mention its name, and I will know which code you are referring to.

\item \user{User:}   Are there any objects that you have not identified so far that must be identified to check the completion of the goal.

\item \gpt{VLM:} Identify the red cubes using the "o1\_ID\_script".
Identify the green containers using the "o2\_ID\_script".
Once both sets of objects are identified, the check for completion would involve verifying that each red cube is within the bounds of a green container. This can be determined by checking if the centroid of each red cube's bounding rectangle falls within the bounding rectangle of any green container. If all red cubes satisfy this condition, the goal is achieved.

\item \user{User:} How would you know if the goal is done?  Please propose one best guess of the check for completion.  The type of check for completion you use should be implementable using a python script.  Please describe this check. The script can only use the first image I gave you.  You are allowed to compare this to the image from the initial frame.  You can store the location of objects expected to be stationary.

\item \gpt{VLM:} To determine if the goal is completed, a Python script can perform the following check for completion:

Check for Completion:

Identify Red Cubes: Use "o1\_ID\_script" to locate all red cubes in the current frame.
Identify Green Containers: Use "o2\_ID\_script" to locate all green containers in the current frame.
Compare Positions: For each identified red cube, the script should calculate the centroid of its bounding rectangle. Then, for each green container, the script should check if any red cube's centroid is within the bounds of the container's bounding rectangle.
Completion Condition: If all red cube centroids are inside the bounds of any green container, then the goal is considered to be achieved.
The Python script would execute these steps and return True if all red cubes are inside the green containers, and False otherwise.

Here is a high-level description of what the Python script would do:

Load the initial and current images.
Apply "o1\_ID\_script" and "o2\_ID\_script" to identify red cubes and green containers in both images.
Calculate the centroids of the red cubes.
Determine the bounding rectangles of the green containers.
Check if each red cube's centroid from the current frame is inside a green container.
Compare the initial and current positions of the red cubes to confirm they have been moved into the containers.
The script returns True if all red cubes are inside the green containers, False otherwise.
This script assumes that the initial and current images are provided and that the cubes and containers are distinguishable by their colors and shapes. The "o1\_ID\_script" and "o2\_ID\_script" provide the necessary object identification for this check.

\item  \user{User: } Please implement this technique, you will only have access to a single frame at a time. If you would like to access the initial state and compare it to the current state or store information from the initial state, this is allowed. Please use the same “o1\_ID\_script” or "o2\_ID\_script” as part of your implementation. You can only use the first image I gave you as input to this script.  This script must return \{True, False\}. It may be useful to store the location of expected static objects from the initial image.  I will show you two images, the first image is of the initial state and the 2nd image is after the goal completion.  You must return False on the first image and true on the 2nd. A few useful techniques: Use the contours to determine if an object is inside of another, don’t approximate rectangles or radii.  Try to approximate shapes in contours as they may not be exact (or incomplete). You can do this by finding a convex hull of every contour which would always be closed and then approximating it down. Checking the shape of objects is useful once you have identified them by color. This script should not require input from me.  

\item \gpt{VLM: } The script correctly identified that the goal has not been completed in the initial state image, returning False. This script can now be applied to a second image, which represents the state after the goal completion, to verify if it returns True as expected for the completed state. Please provide the second image to proceed with the verification. 

\item \user{User: } Provide Final Image

\item \gpt{VLM: } The script has analyzed the final state image and determined that the goal of placing red cubes into green containers has been completed, returning True. This indicates that the conditions for the goal have been met in the provided image. 

\item \user{User: } Can you also propose a real valued reward function that is incremental if possible?

\item \gpt{VLM: } Certainly, in reinforcement learning and many task completion scenarios, a reward function provides incremental feedback to an agent based on its actions. For the task of placing red cubes into green containers, an incremental reward function could be designed to provide positive feedback as the agent makes progress towards the goal, and possibly negative feedback for incorrect actions.

Here's a proposal for a real-valued reward function:

Closeness to Container Reward:

For each red cube, calculate the Euclidean distance to the nearest green container.

Normalize this distance by the diagonal length of the image to get a value between 0 and 1.

Invert the normalized distance (1 - normalized distance) to get a reward value where closer cubes have higher rewards.
Sum these values for all red cubes.

Correct Placement Reward:

Assign a fixed reward value for each red cube that is correctly placed inside a green container.

This value should be significant to reflect the completion of the task for each cube.

Penalty for Incorrect Placement:

If a red cube is placed outside any green container, apply a penalty. This could be a negative value or the subtraction of a potential reward.
Completion Bonus:

If all red cubes are inside green containers, provide a large completion bonus.

\end{itemize}

\textbf{Example failed response:}

It appears that the script did not identify any red cubes in the image. This could be due to several reasons such as the color range for red not being accurately defined for the lighting conditions in the image, or the size threshold for detecting cubes not being appropriate.

To resolve this, we would need to fine-tune the color detection thresholds and possibly adjust the size criteria for identifying the cubes. Unfortunately, due to the constraints of this platform, I can't visually inspect the output to make these adjustments, but in a typical development environment, you would iterate on the color range and size parameters while visually inspecting the output until the red cubes are correctly identified

\section{Verification Pipeline Notes}
We first note that $p=0.1$ in all experiments.

We wish to emphasize the usefulness of the verification pipeline given a few observed examples.

In one suggested completion check for obtaining the key task during the MiniGrid experiments, the VLM used the agent proximity, with a Euclidean distance threshold, to the key as an indication that the reward function.  The agent came into close enough proximity to the key in 43\% of the random trajectories, which was greater than our 10\% threshold.  This was obviously not a useful check for completion and resulted in a more fine-tuned check with refinement.  The final completion check was to verify that the key had disappeared from the image (and was obtained by the agent.  

In another example the agent identified the blue blocks incorrectly in the separating-piles task.  All the blue blocks were identified as a single contour with the center calculated incorrectly.  This center was found to be inside the yellow square over 10\% of the time and therefore the script was not verified.

We also observed another interesting caveat of our verification pipeline, which we discuss.  In CLIPort it is possible that the VLM only identifies one red block correctly.  We did not see this behaviour in our experiments but we wish to note it as a potential problem in other environments.  If one red object is identified, the entire verification pipeline will succeed given the programs are correct in checking the placement into the green bowl.  We therefore suggest when using our method that we have an additional completion check.  The final goal should be obtained within $x$ frames of the trajectory completion to prevent this scenario from happening.

\section{Experiment Parameters}
We list parameters for each of the environment experiments in Tables \ref{miniparams} and \ref{pdgym}

\begin{table*}[t]
\caption{The hyper-parameters used for PPO in the the MiniGrid experiments}
\label{miniparams}
\vskip 0.15in
\begin{center}
\begin{small}
\begin{sc}
\begin{tabular}{lccr}
\toprule
      Parameter & Value  \\
      \midrule
      Adam optimizer $\alpha$                                 & $7\cdot 10^{-4}$      \\ $\beta_1 ; \beta_2$ & $ 0.9 ; 0.999$ \\ $\epsilon$ & $10^{-5}$\\
      entropy coefficient
      & $10^{-2}$\\
      value loss coefficient & 0.5\\
            $\gamma$                                        & 0.99      \\

      maximum norm of gradient &0.5\\
      Time-steps &$1e6$\\
      Batch Size &256\\
      entropy coefficient                    & $10^{-2}$      \\
      clip parameter &0.2
             \\
\bottomrule
\label{tab}
\end{tabular}
\end{sc}
\end{small}
\end{center}
\vskip -0.1in
\end{table*}

\begin{table*}[t]
\caption{The hyper-parameters used for TQC in the the PandasGym experiments}
\label{pdgym}
\vskip 0.15in
\begin{center}
\begin{small}
\begin{sc}
\begin{tabular}{lccr}
\toprule
Parameter & Value  \\
\midrule
  Buffer Size & $1e6$\\
  Batch Size & $2048$\\
     $\gamma$ & $0.95$\\
    $\alpha$ &  $0.001$      \\
      Time-steps & $100,000$\\
       Policy net & $[512, 512, 512]$ \\
       $n$-Critics & $2$   \\
Replay Buffer   &  HER Buffer\\
        Goal Selection Strategy & Future \\
         $n$-sampled goals & 4 \\
         $\tau$ & 0.05 \\
\bottomrule
\label{tab}
\end{tabular}
\end{sc}
\end{small}
\end{center}
\end{table*}
\section{Environment Design}
We make changes to our environments for more efficient use with our pipeline.

\textbf{MiniGrid:} All doors and keys are set to yellow.  The goal is always a green square and agent is always a red arrow.  The reward is $1$ if the task is completed and $0$ otherwise.

\textbf{CLIPort:} The blocks, yellow square, and bowls are always the same colour. 

\textbf{Pandas-Gym:} The block is always green, the goal area is always yellow.  The camera is placed at the front of the table looking downwards at a 45 degree angle towards the robot and objects on the table. The rewards are always sparse, not dense.

\section{Generated Scripts}
All generated scripts are provided in the Supplementary Material.  We name the functions with the corresponding environment name. GPT-4 generated comments were removed.

\end{document}